%% file: main.tex
\documentclass[conference]{IEEEtran}
\IEEEoverridecommandlockouts
\usepackage[utf8]{inputenc}
\usepackage{balance}
\usepackage{subfig}
\usepackage{cite}
\usepackage{amsmath,amssymb,amsfonts}
\usepackage{graphicx}
\usepackage{textcomp}
\usepackage{adjustbox}
\usepackage{xcolor}
\usepackage{mathtools,lipsum, nccmath,cuted}
\usepackage{siunitx}
\usepackage{textcomp}

\usepackage{comment}
\usepackage{makecell}
\usepackage{algorithm}
\usepackage{algorithmic}
\usepackage{gensymb}
\usepackage[normalem]{ulem}
\usepackage{colortbl}
\usepackage{tabularx}
\usepackage{array}
\usepackage{ulem}
\usepackage{caption}
\usepackage[textsize=tiny, textwidth=1.3cm,disable]{todonotes}
\newcommand{\ad}[1]{\todo[color=yellow!50, linecolor=black!50]{\textbf{Abhishek}: #1}}

\usepackage{enumitem}
\usepackage{soul}
\usepackage[capitalise]{cleveref}
\usepackage{multirow}
\author{\IEEEauthorblockN{Vijaya Kumar Sundar \IEEEauthorrefmark{2}\textsuperscript{\textsection}, Shreyas Ramakrishna\IEEEauthorrefmark{1}\textsuperscript{\textsection}, Zahra Rahiminasab\IEEEauthorrefmark{2}, Arvind Easwaran \IEEEauthorrefmark{2}, Abhishek Dubey\IEEEauthorrefmark{1}
}

\IEEEauthorblockA{\IEEEauthorrefmark{2}
\textit{Nanyang Technological University}
}
\IEEEauthorblockA{\IEEEauthorrefmark{1}
\textit{Vanderbilt University}
}
}

\begin{document}

%\section{*}{empty}
%\newpage
%\setcounter{page}{0}

\title{Out-of-Distribution Detection in Multi-Label Datasets using Latent Space of $\beta$-VAE}

% \title{Out-of-Distribution Detection using Latent Representations of $\beta$-VAE for datasets with insufficient labels}

\maketitle

\begingroup\renewcommand\thefootnote{\textsection}
\footnotetext{The authors have equally contributed towards this work}

\pagestyle{plain}

\input{abstract.tex}
\input{paper.tex}

% \balance

\bibliographystyle{IEEEtran}
\bibliography{references/references.bib}
\end{document}

%% file: abstract.tex
\begin{abstract}
Learning Enabled Components (LECs) are widely being used in a variety of perception based autonomy tasks like image segmentation, object detection, end-to-end driving, etc. These components are trained with large image datasets with multimodal factors like weather conditions, time-of-day, traffic-density, etc. The LECs learn from these factors during training, and while testing if there is variation in any of these factors, the components get confused resulting in low confidence predictions. The images with factors not seen during training is commonly referred to as Out-of-Distribution (OOD). For safe autonomy it is important to identify the OOD images, so that a suitable mitigation strategy can be performed. Classical one-class classifiers like SVM and SVDD are used to perform OOD detection. However, the multiple labels attached to the images in these datasets, restricts the direct application of these techniques. We address this problem using the latent space of the $\beta$-Variational Autoencoder ($\beta$-VAE). We use the fact that compact latent space generated by an appropriately selected $\beta$-VAE will encode the information about these factors in a few latent variables, and that can be used for computationally inexpensive detection. We evaluate our approach on the nuScenes dataset, and our results shows the latent space of $\beta$-VAE is sensitive to encode changes in the values of the generative factor. 

\end{abstract}

%% file: paper.tex
\section{Introduction}
\label{sec:intro}

{\bf Emerging Trends:} Cyber-Physical Systems (CPS) are heavily relying on the use of Learning Enabled Components (LECs) \cite{tuncali2018reasoning} to achieve higher levels of autonomy. Specially, in autonomous vehicles, LECs based on perception have become very prominent to perform a variety of perception and control tasks like image segmentation, object detection and end-to-end learning. These LECs are usually trained with large multi-label datasets (e.g. nuScenes \cite{caesar2019nuscenes}) containing images with various multimodal generative factor like lighting (day, night), weather (fog, rain), traffic density, etc. The LECs learn from these common factors and performs exceedingly well if they remain same during testing. However, if the test images have a variation in these factors, then the LECs get confused resulting in low confidence predictions. The images with factors not seen during training is commonly referred to as Out-of-Distribution (OOD). For safe autonomy \cite{neema2019assured} it is important to detect the OOD images, so that a safe mitigation strategy could be performed.

{\bf State of the art:} The problem of OOD detection for multi-label datasets is often transformed into several one-class OOD detection problem. The authors in \cite{vyas2018out} address the similar problem by synthesizing the dataset into mutually exclusive partitions and then train an ensemble of one-class classifiers for OOD detection. One-class classifiers using classical kernel-based methods \cite{kwon2005kernel} and one-class SVM \cite{scholkopf2001estimating} have also been used for OOD detection. However, their inefficiency in operating on high-dimensional images has resulted in growing interests towards deep-learning models. An example of a deep learning model is the Deep Support Vector Data Description (Deep SVDD), which has performed exceptionally well as one-class OOD detectors \cite{ruff2018deep}. It draws a compact hyper-sphere enclosing all the training samples, and the OOD samples will be mapped out of this hyper-sphere. Then the distance of the samples from the center of the hypersphere is used as the metric for OOD detection.    

Another widely used deep learning technique uses generative models such as Autoencoders (AE) \cite{denouden2018improving} and Variational Autoencoder (VAE) \cite{an2015variational}. These models use encoder-decoder neural networks to learn the compression and decompression of the training samples. In the process, it provides a compressed latent representation space that encodes information of all the generative factors in the training samples. VAE encodes the latent space as distributions \cite{an2015variational} of generative factors, while the AE just encodes the latent space as a deterministic mapping of the inputs to outputs. The fact that these models can reconstruct the normal samples well and not the OOD samples is exploited for OOD detection.

\begin{figure}[t]
     \centering
     \includegraphics[width=0.8\columnwidth]{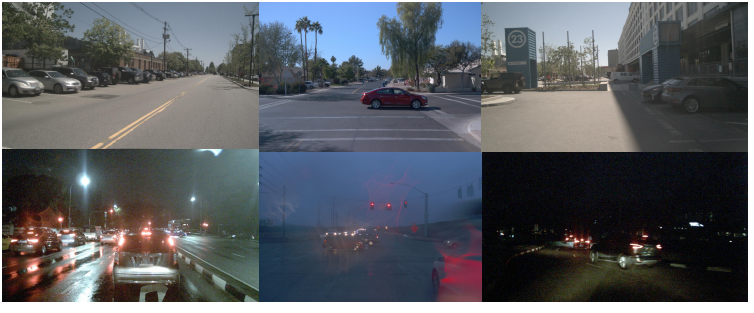}
     \caption{Images from different scenes of the nuScenes dataset \cite{caesar2019nuscenes}. These image have multiple generative factors like time-of-day, weather, pedestrians, traffic, etc. For example, the top right corner image has multiple labels of clear weather, morning, low-traffic, no-pedestrian.}
     \label{fig:nuscenes}
    \vspace{-0.8em}
 \end{figure}
 
\begin{figure*}[t]
    \centering
    \footnotesize
    \includegraphics[width=0.85\textwidth]{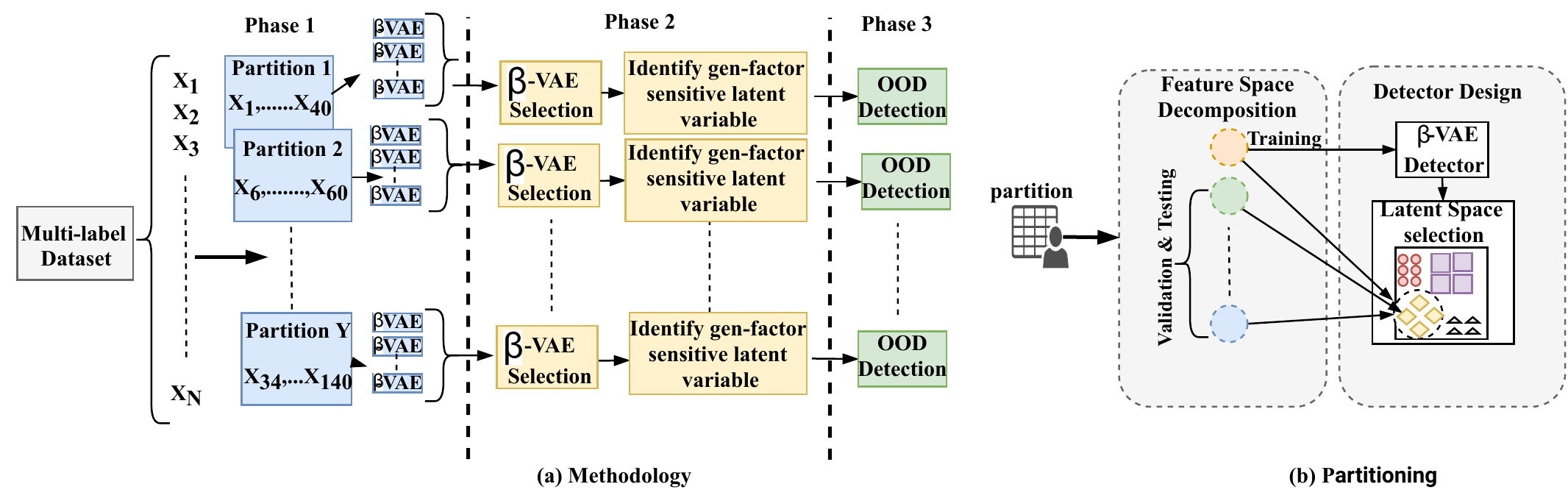}
    \caption{(a) the proposed three-phase methodology for OOD detection. Phase-I: Dataset partitioning and Hyperparameter selection. Phase-II: $\beta$-VAE and informative latent variable selection. Phase-III: Online detection of variations in generative factor value, and (b) the data from a partition is decomposed into several subsets based on the generative factor values. A subset with one generative value is used for training the $\beta$-VAE. A subset containing a mix of different generative factor values is used as validation dataset.}
    \label{fig:design_flow}
    \vspace{-0.8em}
\end{figure*}

Existing AE and VAE based OOD detection work can be categorized into reconstruction-based methods and latent space-based methods. Reconstruction based methods involve computation of normalized difference between each pixel value of the original image and the reconstructed image i.e. the reconstruction error or the reconstruction probability \cite{an2015variational}. The main disadvantage of the reconstruction-based methods is that they are more computationally expensive and, in some cases, can lead to the wrong prediction due to presence of OOD sample on learned manifold. On the other hand, Latent space-based methods \cite{vasilev1806q} compare the distance between the latent distributions of the test and train images. Different metrics like Euclidean distance, Bhattacharyya distance, and Kullback–Leibler (KL) divergence have been used.  

{\bf Research Gap:} A major problem is that these methods have typically been applied to either single-label datasets or datasets with clear partitions (with each partition having mutually exclusive labels). However, the real-world autonomous driving datasets (e.g. nuScenes \cite{caesar2019nuscenes}) are multi-labelled, and synthesizing it into clear partitions based on labels is not possible. For example, each image of nuScenes dataset has multiple labels like time-of-day, weather, pedestrians, traffic-density, vehicle types, etc. Based on these labels, the dataset can only be categorized into approximate partitions, such that each partition has labels of one generative factor fixed, while the labels of other generative factors still change. In such approximate partitions, none of the discussed AE-based OOD detection methods will be able to detect changes in specific generative factors \cite{higgins2017beta}. This is because these OOD detection methods (especially Deep SVDD) can only work as one-class classifiers to learn all the images in the partition, while rejecting the images from the other partitions as OOD.

{\bf Our Contributions:} We propose a three-phase methodology to synthesize the multi-label dataset into smaller partitions and then train a $\beta$-VAE \cite{higgins2017beta} for each partition to generate a compact latent space for OOD detection. $\beta$-VAE is a classical VAE with the $\beta$ hyperparameter, that balances the reconstruction and information channel capacity. Selecting appropriate $\beta>1$, provides $\beta$-VAE the capability to generate a disentangled latent space of the generative factors in the data. This means, that it is possible to identify a single latent variable in the latent space which is sensitive to changes in a specific generative factor, even when the data comes from a partition with multiple labels. The disentangled latent space along with quantitative metrics like KL-divergence and Mean Square Error (MSE) are used in our methodology.  

Briefly our method works as follows. We synthesize the training dataset into partitions based on imprecise labels of image generative factors, such that one generative factor in the partition is fixed, while the others may vary. We then train a $\beta$-VAE for this partition. The $\beta$-VAE is trained as a one-class classifier to learn information only about the factor that was fixed in the partition. Then during testing, we query each of the trained $\beta$-VAE detectors to identify changes in the generative factor value it is was trained with. This approach requires less computational resource compared to the other OOD detection methods we mentioned. For example, SVDD requires a larger latent dimensional space (1024, or 2048) to perform one-class OOD detection. The other benefit is that we can tune the sensitivity of the detectors by adjusting the $\beta$ parameter. In the next section we discuss the methodology to design and use the proposed $\beta$-VAE OOD detector.

\section{Proposed Methodology}
\label{sec:methodology}
In this section, we discuss a three-phase methodology (see \cref{fig:design_flow}) to select a $\beta$-VAE that is sensitive to changes in the values of a specific generative factor.

\subsection{Preliminaries}
\label{sec:prelim}
\textbf{$\beta$-VAE} is a generative model that consists of two attached neural networks named image encoder and decoder. The encoder and decoder learn the posterior distribution $q_{\phi}(z|x)$ and likelihood distribution $p_{\theta}(x|z)$ respectively. To train the $\beta$-VAE, a function named Evidence Lower Bound (ELBO) is maximized, which is the lower bound for the likelihood of data. \cref{eqn:elbo} presents the definition of ELBO function. 

\begin{equation}
\small
    \mathcal{L}(\theta,\phi,\beta;x,z) = {\mathbb{E}}_{q_{\phi}(z|x)}[log p_{\theta}(x|z)]-\beta D_{KL}(q_{\phi}(z|x)||p(z))
    \label{eqn:elbo}
\end{equation}

The expression on the right side in \cref{eqn:elbo} denotes the reconstruction likelihood, which guarantees the similarity between the reconstructed image and the input image. The regularization term in the right hand side of the equation ensures that the distribution learned by $\beta$-VAE is similar to a predefined distribution. It consists of the KL-divergence between the predefined reference distribution of latent variables ($P(z)$), which is fixed as a Gaussian distribution with $\mu$=$0$, $\sigma$=$1$) and the distribution learned by encoder segment. 

\textbf{Disentanglement} as defined in \cite{higgins2017beta}, means that each latent variable mainly encodes the information related to a specific generative factor, and any perturbation of that generative factor will result in significant changes to the values of that latent variable. A  $\beta$-VAE has two hyperparameters, number of latent variables (nLatent) and $\beta$, whose values need to be optimally tuned to obtain a disentangled representation of the latent space. There is no straightforward recipe for finding optimal hyperparameters. However, results from a previous work \cite{higgins2017beta} has shown that $\beta>1$ can better disentangle latent space. An important thing to note in this discussion is, disentanglement can be achieved only if the generative factors in the data are independent. Real world datasets (e.g. nuScenes \cite{caesar2019nuscenes}) may not have independent generative factors, so achieving disentanglement is not possible. So, we try to factorize the latent space to identify the informative latent variables and perform detection using them.

\subsection{Phase-I: Dataset partitioning and Hyperparameter selection}

As the first pre-processing step, we list all possible image generative factors in the dataset. Then, we divide the dataset of images $\{x_1,x_2, .......,x_N\}$ into  partitions\footnote{Note that these are not clean partition and contain labels from other generative factors as well.} based on the  generative factors, such that images belonging to a partition comprises of a specific range of values of the generative factors and are specific to the case study and the dataset considered (details are presented in \cref{sec:experiments}). As shown in \cref{fig:design_flow}-b, we split the partition data into training, validation and testing subsets. The training set has images of one generative factor value (e.g. day). The validation set has images from all the generative factor values (e.g. day, night, evening). The test set has a mix of images from all values of generative factors. However, the test images are not used during training or validation.   

As discussed in \ref{sec:prelim}, $\beta$ and nLatent are the two hyperparameters of $\beta$-VAE that needs to be selected. Selecting the right parameters is very important to obtain a disentangled latent space, but finding optimal values is still an unsolved problem \cite{locatello2018challenging}. However, as suggested in \cite{locatello2018challenging}, we use manual hyperparameter search  \cite{bergstra2012random} over $\beta$ $\in$ $[b_1,b_2,.......b_l]$ and nLatent $\in$ $[n_1,n_2....,n_l]$ to find an optimal combination which minimizes ELBO (\cref{sec:prelim}). This search returns a set of $\beta$ and nLatent combination which is used to train a set of $\beta$-VAE's.

\subsection{Phase-II: $\beta$-VAE and informative latent variable selection}
\label{sec:phase2}
 
The most important phase of our methodology, has two steps: First, we select a $\beta$-VAE from the list of $\beta$-VAE's trained in phase-I. For this we use the following metrics: (1) {\bf Reconstruction quality}: we use reconstruction error of the validation dataset as a metric to select an appropriate $\beta$-VAE. For this we compute the average mean square error (MSE) for the validation dataset images and then select the $\beta$-VAE which has the largest MSE. We base this selection on the fact that an appropriately selected $\beta$-VAE will poorly reconstruct an image from the other generative factor value, and will provide the largest MSE error, and (2) {\bf Regularization term}: we use the regularization term (\cref{eqn:elbo}) or the average KL-divergence across all the latent variables as the metric for $\beta$-VAE selection. We hypothesize that a combination of the two metrics should be used to select an appropriate $\beta$-VAE.

Next, we identify the latent variables that are sensitive to changes in the generative factor value. As we are designing a chain of detectors for different partitions, we restrict the computational requirement of each detector by using only a single latent variable. For this, we find a single latent variable that shows highest sensitivity to the variations in the factor.
Using higher number of latent variables could improve the detection results, however, it increases the detection time. For identifying the latent variable, we perform the following: step1 -- we compute the KL-divergence of the images in the training dataset. The KL-divergence is computed as the dissimilarity between the generated latent distribution and standard normal ($\mu=0$, $\sigma=1$). The KL-divergence metric is given in \cref{eqn:kl-metric} below. 

 \begin{equation}
\small
KL = D_{KL}(q_{\phi}(z|x)||\mathcal{N}(0,I))
\label{eqn:kl-metric}
\end{equation}

where, $\mathcal{N}(0, I))$ and $q_{\phi}(z_n|x)$ are the normal distribution and the distribution generated by the encoder of a $\beta$-VAE for a latent variable $z_n$ respectively. Here $n$ is the number of latent variables. Step2 -- we compute the KL-divergence of all the images in the validation dataset (similar to step1). Step3 -- we use the KL-divergence values computed in step1 and step2, to calculate an average KL-divergence difference across each latent variable using \cref{eqn:abs}.

\begin{equation}
\small
KL_{diff} = |\frac{1}{N} \sum_{l=1}^{N} {KL}_{l}^t -\frac{1}{N} \sum_{l=1}^{N} {KL}_{l}^v|
\label{eqn:abs}
\end{equation}

where, $KL_{l}^t$ indicates the computed KL-divergence of a latent variable in the training dataset, and $KL_{l}^v$ indicates the computed KL-divergence of a latent variable in the validation dataset. The $N$ is the number of samples in the dataset. We compute this difference independently across all the latent variables (z). We then choose the variable with the largest KL-divergence difference to be the variable encoding information about that generative factor.

Then, we choose the threshold ($\tau$) for OOD detection such that, the KL-divergence of the majority of the training samples lie below $\tau$ and the majority of the validation dataset lie above $\tau$. As a heuristic, we set this threshold to at least be above the 70th percentile of the training sample KL-divergence values. This value is case study specific, any $\tau$ that can clearly differentiate between the train and validation samples KL-divergence value should be chosen. 

\begin{figure*}[t]
\setlength{\abovecaptionskip}{-2pt}
     \centering
     \includegraphics[width=\textwidth]{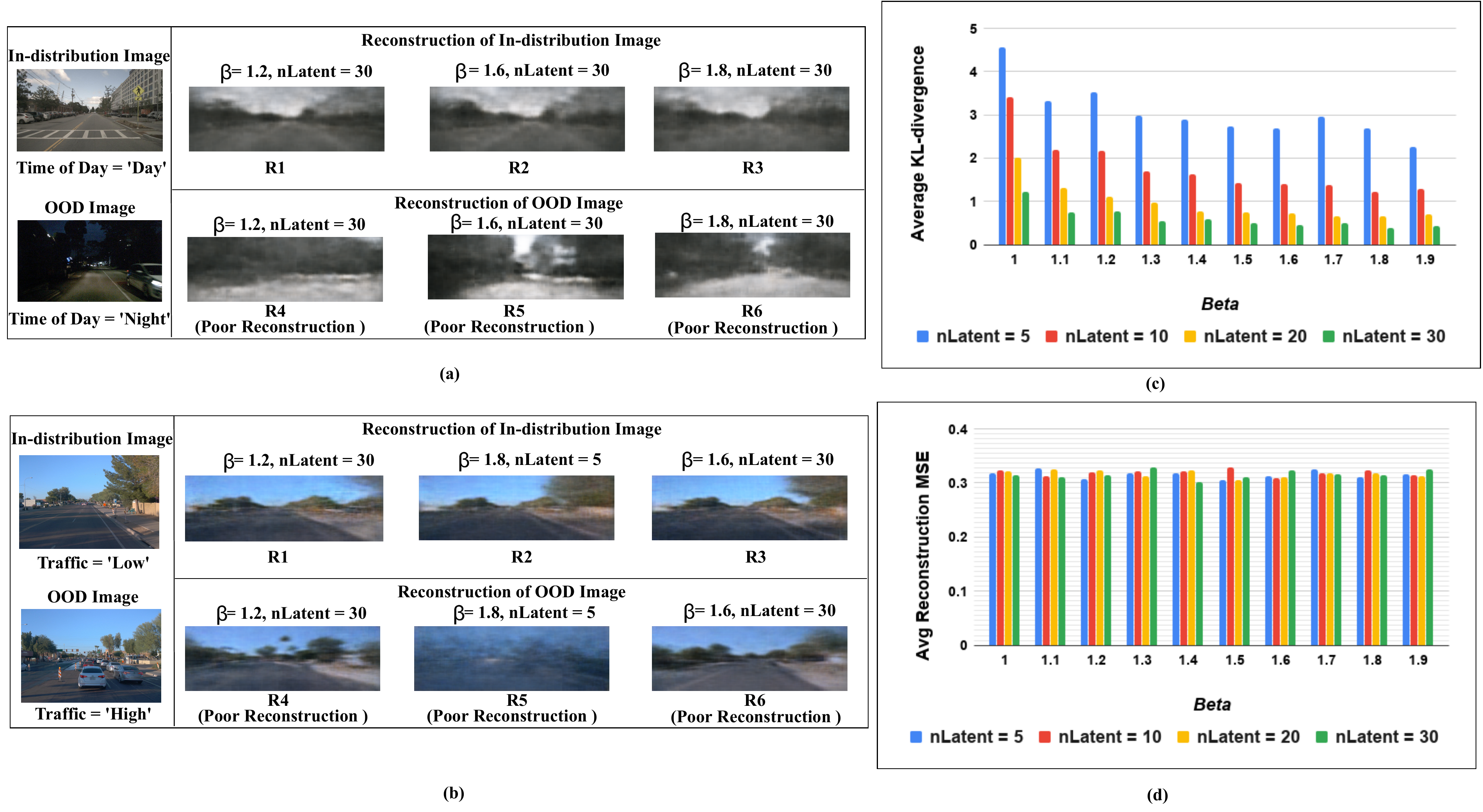}
     \caption{(a) reconstruction of day and night images by different $\beta$-VAEs. For all the three $\beta$ and nLatent combinations, the night images are reconstructed poorly. From this, we hypothesize that all three $\beta$-VAEs are sensitive to the time-of-day generative factor. (b) reconstruction of low-traffic and high-traffic images by different $\beta$-VAEs. For $\beta$=1.8 and nLatent=5, the $\beta$-VAE reconstructs the low-traffic images successfully, while poorly reconstructing the high-traffic images. For other combinations, the reconstruction of high-traffic is similar to the low-traffic images. (c) the average KL-divergence for different $\beta$ and nLatent combinations selected for the time-of-day partition. As seen all the $\beta$ values for nLatent=30 has the lowest KL-divergence compared to the other nLatent. So, we choose nLatent=30 in our experiment. (d) the average reconstruction MSE of validation images for different $\beta$ and nLatent combinations for the time-of-day partition. We use the average MSE as a metric for $\beta$-VAE selection. For the time-of-day partition, all the $\beta$-VAEs reconstruct with a similar MSE, indicating all of them are capable of encoding information about time-of-day. }
     \label{fig:time-of-day}
    \vspace{-0.8em}
 \end{figure*}

\subsection{Phase-III: Online detection of variations in generative factor value}
For online detection, we deploy a chain of $\beta$-VAE's, with each one detecting variation in a specific generative factor value. For example, we design one $\beta$-VAE each to detect variations in the specific value of weather conditions, lighting levels, time-of-day, etc. In this setup, the test images are passed through the chain of $\beta$-VAE's, and in each $\beta$-VAE we compute the KL-divergence (using \cref{eqn:kl-metric}) of the latent variable that was identified in the previous phase. We then compare the KL-divergence value against $\tau$. A value greater than $\tau$ indicates a variation in the generative factor that the $\beta$-VAE was trained to identify. 

\section{Evaluation}
\label{sec:experiments}
We evaluated our methodology on nuScenes dataset \cite{caesar2019nuscenes}. We run all experiments on a test machine with an AMD Ryzen Threadripper 16-core processor and 4 GPUs. 

\subsection{Phase-I: Dataset partitioning and Hyperparameter selection}
\textbf{Dataset}: The nuScenes dataset used for evaluation comprises of $1000$ scenes, each $20$s long and fully annotated with 3D bounding boxes for $23$ object classes and $8$ attributes. Each scene is provided with a description that conveys additional information regarding the time-of-day, weather, traffic, road type, scenery, rain, pedestrians, night lights, and the vehicle type. Using this information, we decompose the dataset to partitions, such that each partition has images of a specific generative factor irrespective of the others. 

For our experiments, we created three partitions based on the factors, time-of-day, traffic, and pedestrians. The time-of-day partition has images from both day and night scenes. The traffic partition has images from scenes containing low and high number of vehicles. The pedestrian partition has images from scenes containing low and high number of pedestrians. The training dataset from each partition is curated to have images from one of the generative factor values (e.g. day). The validation dataset has images from all the generative factors values (e.g. day and night). For testing we prepare two test-sets: test-set1 has images from same scene with same generative factor value (e.g. day), test-set2 has images from multiple scenes but with different generative factor value (e.g. night). The test-set2 images are not used during training.

\textbf{Partitions}: For the time-of-day partition, the training dataset consists of 1000 day images, irrespective of the values of the other generative factors. Test-set1 contains 100 day images that is not seen during training. Test-set2 contains 100 night images from multiple scenes. The pedestrian partition has the same training and test datasets as the traffic partition. This is important to show how detectors sensitive to different labels can be designed using the same data partition. The training dataset for these partitions consists of 1000 low-traffic images without pedestrians. Test-set1 contains 100 low-traffic and no pedestrian images that is not seen during training. Test-set2 contains 100 images of high-traffic with pedestrians. For these partitions images of heavy traffic with pedestrians are considered as OOD images.

\textbf{Hyperparameter selection}: We performed a manual search over a list of $\beta$ $\in$ [1.0, 1.9], varying it in steps of 0.1 and nLatent $\in$ [5, 10, 20, 30] to select a set of $\beta$-VAE's that resulted in loss lower than 1 x $e^{-6}$. For this we designed a $\beta$-VAE that was based on the NVIDIA DAVE-II \cite{bojarski2016end} CNN. The encoder network has three convolutional layers 24/36/48 with (5x5) filters and two convolutional layers 64/64 with (3x3) filters and four fully connected layers with 1164,100, 50 and 40 units. A symmetrical decoder architecture is designed as the other end. We then train the model along with the hyperparameters for 100 epochs at learning rate = 1 x $10^{-5}$, using adam optimizer. The search returned a set of $\beta$-VAE's with low ELBO loss. For all three partitions, several $\beta$ values with nLatent=30 returned the lowest ELBO and average KL-divergence loss. The average KL-divergence loss across different hyperparamter combination is shown in \cref{fig:time-of-day}-c.

\subsection{Phase-II: $\beta$-VAE and informative latent variable selection}
 
\textbf{Time-of-day partition}: \cref{fig:time-of-day}-a shows the images reconstructed by different $\beta$-VAE's. Images R1 to R6 represent reconstruction of day images while images R7 to R12 shows the reconstruction of night images. We observe that all the three $\beta$-VAE's reconstruct the day images with a reasonable quality and the night images with poor quality. Values of the average reconstruction mean square error (MSE) is shown in \cref{fig:time-of-day}-d. The average MSE is very similar for all the combinations of $\beta$-VAE's, indicating all the three $\beta$-VAE's are sensitive to the changes in the time-of-day factor. Therefore to choose one $\beta$-VAE among the three we compute the average KL-divergence of all the latent variables.

\cref{fig:time-of-day}-c shows the average KL-divergence loss for the different combinations of $\beta$-VAE. From this we found $\beta$=1.8, and nLatent=30 combination resulted in the lowest KL-divergence loss, so we selected it. We then identified V6 (the 6th latent variable of the 30 latent variables) to be most sensitive to variation in time-of-day value, so selected it for OOD detection.

% \ad{explain that this means the 6th latent variable out of -- how many}

\textbf{Traffic partition}: \cref{fig:time-of-day}-b illustrates the image reconstructions by different $\beta$-VAE's for the traffic partition. In \cref{fig:time-of-day}-b, R1 to R3 represents reconstruction of low-traffic images while R4 to R6 shows the reconstruction of high traffic images. The $\beta$-VAE with $\beta$=1.8, and $nLatent=5$, reconstructs the high traffic images poorly compared to the other combinations. So, we consider $\beta$-VAE trained with this specific combination to be highly sensitive to changes in traffic. For the selected combination of $\beta$-VAE hyperparameters, we found latent variable V15 (the 15th latent variable of the 30 latent variables) to be the most sensitive to variation in traffic information. 

\textbf{Pedestrian partition}: Selecting a $\beta$-VAE for encoding information about pedestrians was challenging, as none of the $\beta$-VAE could successfully reconstruct the pedestrian information. None of the $\beta$-VAE's trained in phase-I were sensitive enough to capture variations in the values of pedestrian factor. So, we selected the $\beta$-VAE based on the average KL-divergence loss. Similar to the other partitions we found latent variable V30 (the 30th latent variable of the 30 latent variables) to be sensitive to variation in pedestrian information.

\begin{figure}[t]
     \centering
     \includegraphics[width=0.96\columnwidth]{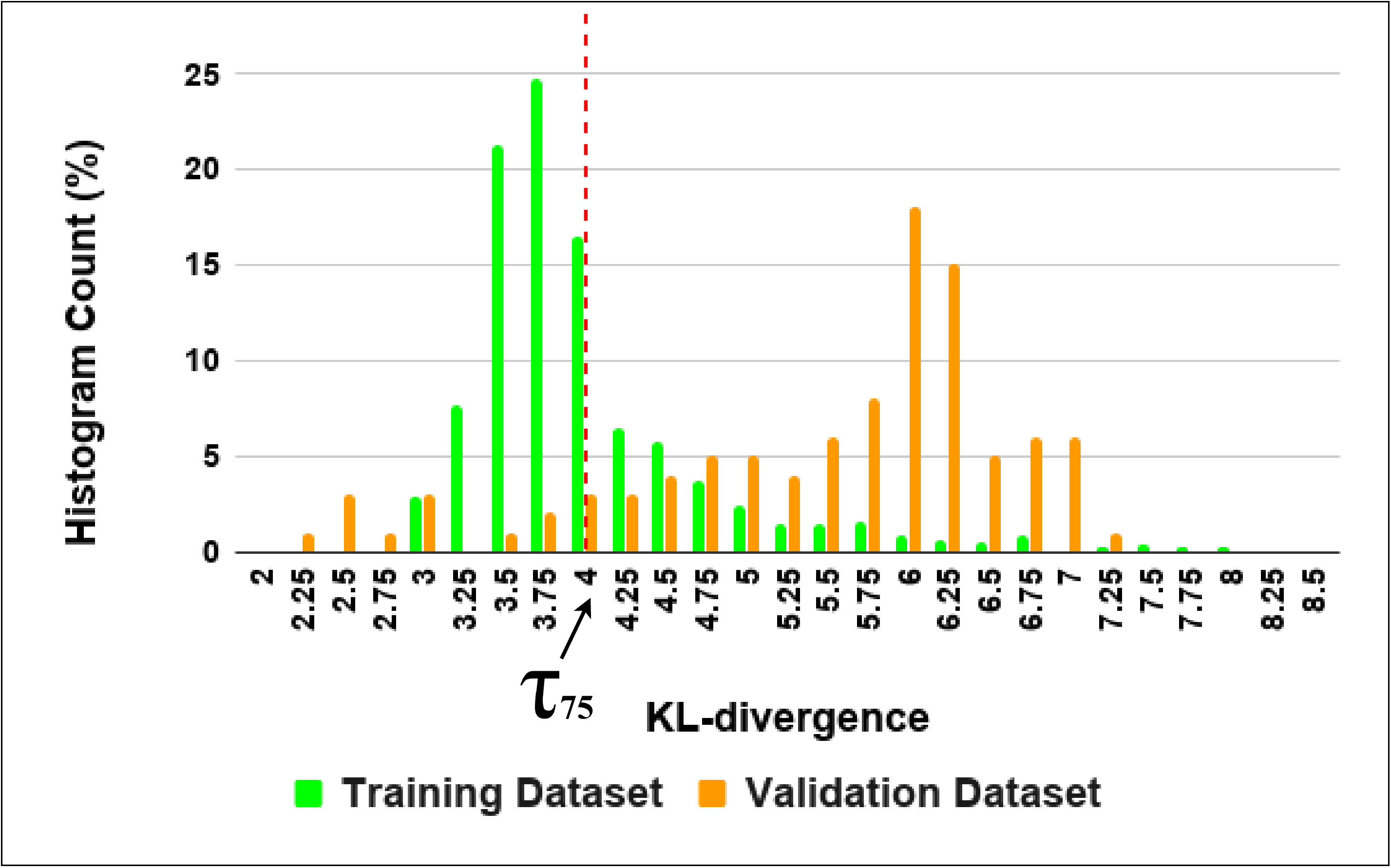}
     \caption{Selecting threshold using the training and validation datasets of time-of-day partition. We set the threshold $\tau_{75}$ at the 75th percentile of the KL-divergence value. Any test sample with KL-divergence value $>$ $\tau_{75}$ will indicate the value of generative factor has changed.}
     \label{fig:hist}
    \vspace{-0.8em}
 \end{figure}
 
\input{Tables/newkltable.tex}

Next, we find $\tau$ for each of the partition. As discussed in \cref{sec:phase2} we select $\tau$ as the percentile of the training samples. \cref{fig:hist} shows the $\tau$ selection for time-of-day partition. Based on our experiments we found $\tau_{75}$ as the optimal threshold to avoid high false positives. \cref{table:resulttable} summarizes the $\tau$ values of each partition.

\subsection{Phase-III: Online detection of variation in generative factor value}
During online detection, the test images were passed through all the three $\beta$-VAE's in parallel. In each $\beta$-VAE detector, the KL-divergence of the test image latent distributions was computed and compared against the identified $\tau$. A KL-divergence value was greater than $\tau$ indicated the generative factor value of the test image had changed.  \cref{table:resulttable} summarizes some key detection results from the three partitions. For compiling the performance of the detectors, we compared the outcome of the detector to the actual generative factor label. For example, for time-of-day partition if the actual image label was day, we compare this against the detector result.

For the time-of-day partition, the $\beta$-VAE detector with ($\beta$=1.8 and nLatent=30) identified 99\% of the images from test-set1 to be in-distribution, while it correctly identified 95\% of images from test-set2 to be OOD. For the traffic partition, the $\beta$-VAE detector with ($\beta$=1.6 and nLatent=30) identified 90\% of the images from test-set1 to be in-distribution, while it correctly identified 74\% of images from test-set2 to be OOD. Also, for the pedestrian partition, the detector identified 100\% of the images from test-set1 to be in-distribution, while it correctly identified 100\% of images from test-set2 to be OOD.

\ad{can you say anything about the resource usage -- that was one of the big factors}

\subsection{Discussion and Future Work}
Our evaluations show that $\beta$-VAE detectors can reliably identify variations in the values of the specific generative factor with very few false positives. The OOD detection results depend on the threshold selection. Our threshold selection criteria aims at reducing the false positives, so we try to find a threshold value that encompasses most of the training data samples. In addition, the detection results were also influenced by the complexity of the scene and the generative factor of interest. The time-of-day factor was simpler among all three factors as its variations is evident and could be encoded by any of the $\beta$-VAE combination. However, identifying variations in the traffic factor and the pedestrian factor was complicated as their variations were much subtle compared to time-of-day. Detecting variation in these factors was scene specific. For scenes with too much variation in the background information, these $\beta$-VAE's did not work reliably. For these scenes, the background information got encoded in all the latent variables and finding a latent variable encoding information about traffic or pedestrian was difficult. In this case, even increasing nLatent to a large value (e.g. 100) did not work. 

Currently, we use minimum ELBO as our hyperparameter selection criteria, but this does not guarantee the selection of optimal hyperparameters as explained in \cite{locatello2018challenging}. To address this we are working towards using a better selection criteria using a disentanglement metric like the BetaVAE metric \cite{higgins2017beta}. Also, we are currently using a single threshold value for OOD detection, which results in high false positives, to overcome this we are working on incorporating change point techniques over a time window to consider a series of image for OOD detection.  

\section{Conclusion}
\label{sec:conclusion}
In this work, we discuss the problem of detecting changes in the generative factor of images in large multimodal autonomous driving datasets. As discussed, these datasets cannot be clearly partitioned, thus making it difficult to apply the existing OOD detection methods. We address this problem using the disentangled latent spaces of the $\beta$-VAE. For performing OOD detection using $\beta$-VAE, we provide a methodology to: (1) select an appropriate $\beta$-VAE with right disentanglement, and (2) select a latent variable that is sensitive to changes in a specific generative factor. The selected latent variable is then used for OOD detection. We have further illustrated the utility of our methodology on the nuScenes dataset.

\section*{Acknowledgement}
This work was supported in part by DARPA's Assured Autonomy project and Air Force Research Laboratory. Any opinions, findings, and conclusions or recommendations expressed in this material are those of the author(s) and do not necessarily reflect the views of DARPA or AFRL.

%% file: Tables/newkltable.tex
\begin{table}[h!]
\centering
\footnotesize
\begin{tabular}{|l|l|l|l|}
\hline
\multicolumn{1}{|c|}{\textbf{Factor}}                           & \multicolumn{1}{c|}{\textbf{\begin{tabular}[c]{@{}c@{}c@{}}Selected $\beta$, \\ nLatent, \\ latent variable\end{tabular}}}                                                          & \multicolumn{1}{c|}{\textbf{\begin{tabular}[c]{@{}c@{}}KL-divergence \\ Threshold ($\tau$)\end{tabular}}}                                                    & \multicolumn{1}{c|}{\textbf{\begin{tabular}[c]{@{}c@{}}\% Test images \\ detected OOD\end{tabular}}}                                                  \\ \hline
\textbf{\begin{tabular}[c]{@{}c@{}}Time-of-day\end{tabular}} & \begin{tabular}[c]{@{}c@{}}$\beta$ = 1.8,\\ nLatent = 30,\\ V6 \end{tabular}                                                                                  & \begin{tabular}[c]{@{}c@{}} $\tau$=4.0 \\ (75th percentile \\of  training \\ images)\end{tabular}                                                                    & \begin{tabular}[c]{@{}c@{}} Test-set1: 1\% \\Test-set2: 95\%\end{tabular}                                                                             \\ \hline
\textbf{Traffic}                                                & \begin{tabular}[c]{@{}c@{}}$\beta$ = 1.6,\\ nLatent = 30,\\ V15 \end{tabular} & \begin{tabular}[c]{@{}c@{}}$\tau$=1.45 \\ (70th percentile \\of  training \\ images)\end{tabular} & \begin{tabular}[c]{@{}c@{}}Test-set1: 10\% \\Test-set2: 74\%\end{tabular} \\ \hline
\textbf{Pedestrian}                                             & \begin{tabular}[c]{@{}c@{}}$\beta$ = 1.8,\\ nLatent = 30,\\ V30\end{tabular}                                                                                 & \begin{tabular}[c]{@{}c@{}}$\tau$=0.06 \\ (92th percentile \\of  training \\ images)\end{tabular}                                                                                     & \begin{tabular}[c]{@{}c@{}}Test-set1: 0\%\\Test-set2: 100\%\end{tabular}                                                                         \\ \hline
\end{tabular}
\caption{The summary of results from the three partitions. The selected $\beta$-VAE hyperparameters, most informative latent variable, $\tau$ and the \% of test OOD images detected is listed. The test results are reported based on five trail runs.}
\label{table:resulttable}
\setlength{\abovecaptionskip}{-8pt}
\vspace{-0.8em}
\end{table}